\documentclass{article} 
\usepackage{arxiv,times}

\usepackage{amsmath,amsfonts,bm}

\def\eqref#1{equation~\ref{#1}}

\def\floor#1{\lfloor #1 \rfloor}
\def\1{\bm{1}}

\DeclareMathAlphabet{\mathsfit}{\encodingdefault}{\sfdefault}{m}{sl}
\SetMathAlphabet{\mathsfit}{bold}{\encodingdefault}{\sfdefault}{bx}{n}

\usepackage{graphicx}
\graphicspath{ {figures/} }
\usepackage{subcaption}
\usepackage{hyperref}
\usepackage{wrapfig}
\usepackage{multirow}
\usepackage{booktabs}       
\usepackage{url}

\title{Mix \& Match: training convnets with mixed image sizes for improved accuracy, speed and scale resiliency}
\author{Elad Hoffer, Berry Weinstein \& Itay Hubara  \\
Habana Labs Research\\
Caesarea, Israel \\
\texttt{\{ehoffer,bweinstein,ihubara\}@habana.ai}
\And
Tal Ben-Nun \& Torsten Hoefler \\
Department of Computer Science\\
ETH Zurich\\
\texttt{\{talbn,htor\}@inf.ethz.ch}
\AND
Daniel Soudry \\
Department of Electrical Engineering \\
Technion, Haifa, Israel \\
\texttt{daniel.soudry@gmail.com}
}

\iclrfinalcopy 
\begin{document}

\maketitle

\begin{abstract}
Convolutional neural networks (CNNs) are commonly trained using a fixed spatial image size predetermined for a given model. Although trained on images of a specific size, it is well established that CNNs can be used to evaluate a wide range of image sizes at test time, by adjusting the size of intermediate feature maps. \\
In this work, we describe and evaluate a novel mixed-size training regime that mixes several image sizes at training time. We demonstrate that models trained using our method are more resilient to image size changes and generalize well even on small images. This allows faster inference by using smaller images at test time. For instance, we receive a $76.43\%$ top-1 accuracy using ResNet50 with an image size of $160$, which matches the accuracy of the baseline model with $2 \times$ fewer computations.
Furthermore, for a given image size used at test time, we show this method can be exploited either to accelerate training or the final test accuracy. For example, we are able to reach a $79.27\%$ accuracy with a model evaluated at a $288$ spatial size for a relative improvement of $14\%$ over the baseline.
Our PyTorch implementation and pre-trained models are publicly available\footnote{\url{https://github.com/eladhoffer/convNet.pytorch}}.



\end{abstract}

\section{Introduction}

\begin{wrapfigure}{r}{0.5\textwidth}
\vspace{-1em}
  \begin{center}
    \includegraphics[width=0.5\textwidth,trim={0 0 0 0cm},clip]{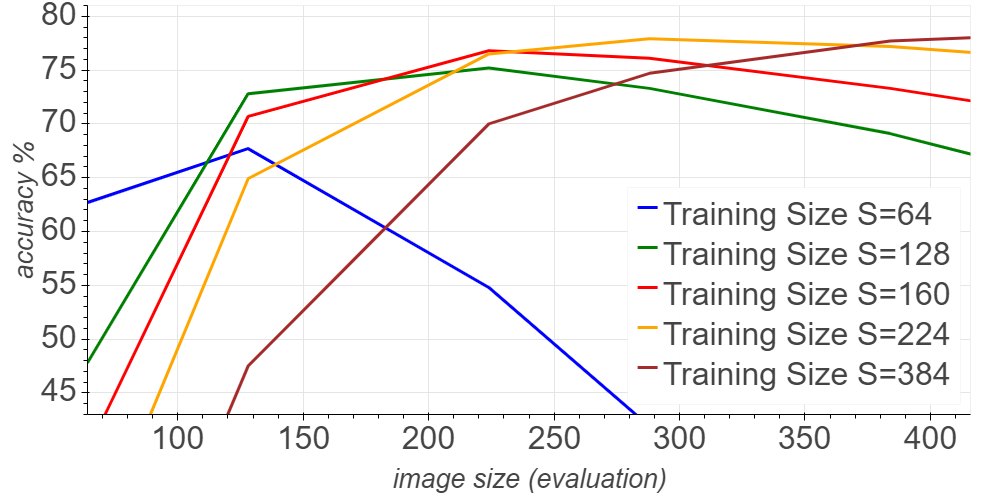}
    \caption{Test accuracy per image size, models trained on specific sizes (ResNet50, ImageNet). }\label{motivation_sizes}
    \end{center}
    \vspace{-2em}
\end{wrapfigure}  
Convolutional neural networks are successfully used to solve various tasks across multiple domains such as visual \citep{krizhevsky2012imagenet, ren2015faster}, audio \citep{van2016wavenet}, language \citep{gehring2017convolutional} and speech \citep{abdel2014convolutional}. 
While scale-invariance is considered important for visual representations \citep{lowe1999object}, convolutional networks are not scale invariant with respect to the spatial resolution of the image input, as a change in image dimension may lead to a non-linear change of their output. Even though CNNs are able to achieve state-of-the-art results in many tasks and domains, their sensitivity to the image size is an inherent deficiency that limits practical use cases and requires evaluation inputs to match training image size. For example, \citet{fixresnet} demonstrated that networks trained on specific image size, perform poorly on other image sizes at evaluation, as shown in Figure \ref{motivation_sizes}. 


Several works attempted to achieve scale invariance by modifying the network structure \citep{xu2014scale, takahashi2017scale}. However, the most common method is to artificially enlarge the dataset using a set of label-preserving transformations also known as "data augmentation" \citep{krizhevsky2012imagenet, howard2013some}. Several of these transformations scale and crop objects appearing within the data, thus increasing the network's robustness to inputs of different scale.

Although not explicitly trained to handle varying image sizes, CNNs are commonly evaluated on multiple scales post training, such as in the case of detection \citep{lin2017focal, yolov3} and segmentation \citep{he2017mask} tasks. In these tasks, a network that was pretrained with fixed image size for classification is used as the backbone of a larger model that is expected to adapt to a wide variety of image sizes.

In this work, we will introduce a novel training regime, ``MixSize" for convolutional networks that uses stochastic image and batch sizes. The main contributions of the MixSize regime are:
\begin{itemize}
    \item \textbf{Reducing image size sensitivity}. We show that the MixSize training regime can improve model performance on a wide range of sizes used at evaluation.
    \item \textbf{Faster inference}. As our mixed-size models can be evaluated at smaller image sizes, we show up to $2 \times$ reduction in computations required at inference to reach the same accuracy as the baseline model.
    \item \textbf{Faster training vs. high accuracy}. We show that reducing the average image size at training leads to a trade-off between the time required to train the model and its final accuracy.
\end{itemize}


\section{Related work}
\subsection{Using multiple image sizes}
Deep convolutional networks are traditionally trained using fixed-size inputs, with spatial dimensions $H\times W$ and a batch size $B$. The network architecture is configured such that the spatial dimensions are reduced through strided pooling or convolutions, with the last classification layer applied on a $1\times 1$ spatial dimension. Modern convolutional networks usually conclude with a final "global" average pooling \citep{lin2013network, szegedy2015going}, that reduces any remaining spatial dimensions with a simple averaging operation. Modifying the spatial size of an input to a convolutional layer by a factor $\gamma$, will yield an output with size scaled by the same factor $\gamma$. This modification does not require any change to the number of parameters of the given convolutional layer, nor its underlying operation. Small changes in the expected size can occur, however, due to padding or strides performed by the layer. It was observed by practitioners and previous works that a network trained on a specific input dimension can still be used at inference using a modified image size to some extent \citep{simonyan2014very}. 
Moreover, evaluating with an image size that is larger than used for training can improve accuracy up to a threshold, after which it quickly deteriorates \citep{fixresnet}.

Recently, \citet{tan2019efficientnet} showed a computational-vs-accuracy trade-off in scaling image size used to train and evaluate with a convolutional network. This finding is consistent with past findings, which demonstrated that training with a larger image size can result in a larger classification error \citep{szegedy2016rethinking,huang2018gpipe}. In addition, previous works explored the notion of ``progressive resizing"  \citep{karras2017progressive, FastAi} --- increasing image size as training progresses to improve model performance and time to convergence.
More recently, \citet{fixresnet} demonstrated that CNNs can be trained using a fixed small image size and fine-tuned post-training to a larger size, with which evaluation will be performed. This procedure reduced the train-test discrepancy caused by the change in image size and allowed faster training time and improved accuracy --- at the cost of additional fine-tuning procedure and additional computations at inference time. 

In this work we will further explore the notion of using multiple image sizes at training, so the CNN performance will be resilient to test time changes in the image size. 

\subsection{Large batch training of deep networks}
Deep neural network training can be distributed across many computational units and devices. The most common distribution method is by "data-parallelism"---computing an average estimate of the gradients using multiple, separably computed data samples. As training NN models is done using batch-SGD method and its variants, scaling this process across more computational devices while maintaining similar utilization for each device inflates the global batch size.

Large batch training is known to affect the generalization capabilities of the networks and to require modification of the regime used for its optimization. While several works claimed that large-batch training leads to an inherent "generalization gap" \citep{keskar2016large}, more recent works demonstrated that this gap is largely caused from an insufficient number of optimization steps performed and can be partly mitigated by hyper-parameter tuning  \citep{hoffer2017train, shallue2018measuring}.  In order to cope with the changes in the training dynamics of the network, several modifications to the optimization procedure have been proposed such as a linear \citep{goyal2017accurate} or a square-root \citep{hoffer2017train} scaling of the learning rate with respect to the batch size growth. 
Other modifications include per-layer gradient scaling schemes \citep{you2017large} and optimizer modifications \citep{ginsburg2019stochastic}. Several works also explored using incremented batch-sizes \citep{smith2018don} in order to decrease the number of training iterations required to reach the desired accuracy.

Recent work by \citet{hoffer2019augment} introduced the notion of "Batch Augmentation" (BA)---increasing the batch size by augmenting several instances of each sample within the same batch. BA aids generalization across a wide variety of models and tasks, with the expense of an increased computational effort per step. A similar method called ``Repeated Augmentation" (RA) was proposed by \citet{berman2019multigrain}. It was also demonstrated that BA may allow to decrease the number of training steps needed to achieve a similar accuracy and also mitigate I/O throughput bottlenecks \citep{choi2019faster}. 
As previous works investigated mostly homogeneous training settings (e.g., using a fixed batch size), an open question still exists on the utility of rapidly varying batch-sizes. We will explore this notion and suggest a new optimizer modification that enables training with multiple varying batch-sizes with limited hyper-parameter tuning.

\section{MixSize: Training with multiple image scales}
 \label{correlation_scales} The traditional practice of training convolutional networks using fixed-size images holds several shortcomings. First, CNNs are commonly evaluated using a different size than that used for training \citep{lin2017focal, yolov3, he2017mask} and it was observed that classification accuracy may degrade above or below a certain size threshold (\citet{fixresnet} and Figure \ref{motivation_sizes}). To remedy these issues, we suggest a stochastic training regime, where image sizes can change in each optimization step. 

\paragraph{Motivation.} In order to motivate our method, we first evaluate the impact of the image size on the training progress of a CNN --- by examining gradient statistics during training\footnote{We used a ResNet-44 model \citep{he2016deep}, trained on the CIFAR10 dataset \citep{krizhevsky2009learning}, whose standard image size is $32\times32$. Measurements are performed on the whole network's gradient vector. Images were sampled in uniform. The smaller $24\times24$ images were down-sampled with bilinear interpolation.}. Specifically, in Table \ref{tab:correlation} we measured the correlation of the gradients across image sizes. We see that gradients computed across different scales of the same image have a strong correlation compared to those obtained across different images. This correlation is especially apparent during the first stages of training and decreases as the model converges. This suggests that the small image gradients can be used as an approximation of the full image gradients, with a smaller computational footprint. 
Therefore, using large images along the entire training process may be sub-optimal in terms of computational resource utilization. More specifically, as the gradients of images of different size are highly correlated at the initial steps of training, it may prove beneficial to sacrifice spatial size in favor of batch size that can be increased. To do so, we suggest the following.

\begin{table}[h]
	\caption{ResNet-44 gradient correlation on CIFAR10.  We measure the Spearman correlation coefficient $\rho$ between different spatial size of random images  $\rho\left(x^{(s_1)},x^{(s_2)}\right)$, as well as non-identical random images of the same size $\rho\left(x^{(s_1)},y^{(s_1)}\right)$. We also compute the variance $V(x)$ for the gradients of each spatial size.}
	\scriptsize
	\label{tab:correlation}
	\centering
	\begin{tabular}{lrrr}
		\toprule
		Measure & \multicolumn{3}{c}{Network State} \\
		\cmidrule{2-4}
				 & Initial & Partially Trained & Fully Trained \\

		\midrule
		Epoch     & $1$ & $50$ & $100$ \\
        Test Accuracy & $55.12\%$ & $87.56\%$ & $92.62\%$
        \\\addlinespace
        \midrule
        $\rho\left(x^{(32)},x^{(24)}\right)$  & $0.2$  & $0.08$ & $0.03$  \\
        $\rho\left(x^{(32)},y^{(32)}\right)$  &  $0.086$  & $0.02$ & $-0.004$ \\
        \midrule
        $V\left(x^{(32)}\right)$              & $1.03e^{-6}$ & $1.44e^{-6}$ & $6.24e^{-7}$ \\
        $V\left(x^{(24)}\right)$              & $1.95e^{-6}$ & $6.34e^{-6}$ & $2.26e^{-5}$ \\
		\bottomrule
	\end{tabular}
	\vspace{-1em}
\end{table}

\paragraph{The MixSize training regime.}
\label{stochastic_sizes}
We suggest "MixSize", a stochastic training regime, where input sizes can vary in each optimization step. In this regime, we modify the spatial dimensions $H,W$ (height and width) of the input image size\footnote{The input image size is changed using bilinear interpolation. The spatial dimensions of all intermediate maps in the CNN are changed accordingly, at the same scale as the input.}, as well as the batch size. The batch size is changed either by the number of samples used, denoted $B$, or the number of batch-augmentations for each sample \citep{hoffer2019augment}, denoted $D$ ("duplicates"). 
To simplify our notation and use-cases, we will follow the common practice of training on square images and use $S=H=W$.
Formally, in the MixSize regime, these sizes can be described as random variables sharing a single discrete distribution
\begin{equation}
   (\hat{S},\hat{B},\hat{D})= \left\{ (S,B,D)_i \textit{ \ w.p. \ } p_i \right\}, 
\end{equation} where $\forall i : p_i \geq 0  $ and  $\sum_i p_i = 1$.

As the computational cost of each training step is approximately proportional to $S^2\cdot B\cdot D$, 
we choose these sizes to reflect an approximately fixed budget for any choice $i$ such that $S_i^2B_iD_i \approx \mathit{Const}$. Thus the computational and memory requirements for each step are constant. 

\paragraph{Benefits and Trade-offs.}
We will demonstrate that using such a MixSize regime can have a positive impact on the resiliency of trained networks to the image size used at evaluation. That is, mixed-size networks will be shown to have better accuracy across a wide range of sizes. This entails a considerable saving in computations needed for inference, especially when using smaller models. Furthermore, given a fixed budget of computational and time resources (per step), we can now modify our regime along spatial and batch axes. We will explore two trade-offs:
\begin{itemize}
    \item \textbf{Decrease number of iterations per epoch} -- by enlarging $B$ at the expense of $S$.
     \item \textbf{Improve generalization per epoch} -- by enlarging $D$ at the expense of $S$.
\end{itemize}

\section{Improved training practices for MixSize}
MixSize regimes continuously change the statistics of the model's inputs, by modifying the image size as well as batch-size. This behavior may require hyper-parameter tuning and may also affect size-dependent layers such as batch normalization \citep{ioffe2015batch}. To easily adapt training regimes to the use of MixSize as well as improve their final performance, we continue to describe two methods we found useful: \textbf{Gradient Smoothing} and \textbf{Batch-norm calibration}.

\subsection{Gradient smoothing\label{sub:GradientSmoothing}}
Training with varying batch and spatial sizes inadvertently leads to a change in the variance of the accumulated gradients. For example, in Table \ref{tab:correlation}, the gradient variance is larger when computed over a small image size (unsurprisingly). This further suggests that the optimization regime should be adapted to smaller spatial sizes, in a manner similar to learning-rate adaptations that are used for large-batch training. This property was explored in previous works concerning large-batch regimes, in which a learning rate modification was suggested to compensate for the variance reduction for larger batch-sizes. Unfortunately, the nature of this modification can vary from task to task or across models \citep{shallue2018measuring}, with solutions such as a square-root scaling \citep{hoffer2017train}, linear scaling \citep{goyal2017accurate} or a fixed norm ratio \citep{you2017large}. Here we suggest changing both the spatial size as well as the batch size, which is also expected to modify the variance of gradients within each step and further complicates the choice of optimal scaling.

Previous works suggested methods to control the gradient norm by gradient normalization \citep{hazan2015beyond} and gradient clipping \citep{pascanu2013difficulty}. These methods explicitly disable or limit the gradient's norm used for each optimization step, but also limit naturally occurring variations in gradient statistics. 
We suggest an alternative solution to previous approaches, which we refer to as "Gradient smoothing". Gradient smoothing mitigates the variability of gradient statistics when image sizes are constantly changing across training. 

We introduce an exponentially moving weighted average of the gradients' norm $\bar{g}_t$ (scalar) which is updated according to $$\bar{g}_t=\alpha \bar{g}_{t-1} + (1-\alpha)g_t$$ where $$g_t=\left\|\frac{\partial E}{\partial w_t}\right\|_2 \ \ \text{and} \ \  \ \bar{g}_{0} = g_0\,.$$ 
We normalize the gradients used for each step by the smoothing coefficient, such that each consecutive step is performed with gradients of similar norm. For example, for the vanilla SGD step, we use a weight update rule of the form
$$w_{t+1} = w_t - \eta  \frac{\bar{g}_t}{g_t}\frac{\partial E}{\partial w_t} \,.$$ 

This running estimate of gradient norm is similar to the optimizer suggested by \citet{ginsburg2019stochastic}, which keeps a per-layer estimate of gradient moments. Gradient smoothing, however, is designed to adapt globally (across all layers) to the batch and spatial size modification and can be used regardless of the optimization method used.

We found gradient smoothing to be mostly beneficial in regimes where multiple varying batch sizes are used. Figure \ref{gs_compare:grad} in the Appendix demonstrates how gradient smoothing reduces the gap between gradient norms of different sizes. Measuring test error on the same model shows a slight advantage for gradient-smoothing (Appendix Figure \ref{gs_compare:acc}).  

\subsection{Batch-norm calibration for varying image sizes}
\label{calibrate_bn}
As demonstrated by \citet{fixresnet}, using a different image size at evaluation may incur a discrepancy between training and evaluation protocols, caused by using different data pre-processing. \citet{fixresnet} 
suggested a post-training procedure, where a network trained on a specific fixed-size is fine-tuned on another size, later used for evaluation. Their solution required $10$s of training epochs, amounting to $1000$s of full forward and back-propagation computations, along with parameter updates for batch-norm and classifier layers.
In contrast, we surmise that for networks trained with mixed-regimes, discrepancy issues mainly arise from the use of the batch-norm layers \citep{ioffe2015batch} and can be solved by targeting them specifically.

Batch-norm layers introduce a discrepancy between training and test evaluations \citep{ioffe2017batch}, as at inference a running estimate of the mean and variance (of training data) are used instead of the actual mean and variance values. This difference is emphasized further in the use of varying image size, as changing the spatial size of an input map can significantly modify the measured variance of that map. While a fine-tuning process per image size can eliminate this discrepancy \citep{fixresnet}, we offer a simpler alternative. For each evaluated size, we calibrate the mean and variance estimates used for that size by computing an average value over a small number of training examples. 
This calibration requires only a few (100s) feed-forward operations with no back-propagation or parameter update and takes only a few seconds on a single GPU.

Interestingly, we highlight the fact that although this process has little or no effect on models trained using a fixed-size input, it does improve our mixed-size models considerably on a wide range of image sizes.
\section{Experiments}
\subsection{MixSize with a fixed image size at test-time: the speed-accuracy trade-off}
\paragraph{CIFAR10/100.}
First, we examine our method using the common visual datasets CIFAR10/100 \citep{krizhevsky2009learning} that consist of $32 \times 32$ color images. We use the ResNet-44 model suggested by \citep{he2016deep}, Wide Resnet WRN-28-10  \citep{Zagoruyko2016WRN} and AmoebaNet \citep{real2019regularized} with their original regime and batch size of $64$. While for ResNet-44 we use the original augmentation protocol, we apply cutout \citep{devries2017improved} and auto-augment policies \citep{cubuk2018autoaugment} on WRN-28-10 and AmoebaNet for both datasets (see Appendix \ref{experiments:cifar} for details). 

As CIFAR datasets are limited in size, we consider the following balanced stochastic regime chosen: 
  \begin{equation*}
    S=
    \begin{cases}
      40, & \text{w.p. \ } \ p=0.2 \\
      32, & \text{w.p. \ } \ p=0.3 \\
      24, & \text{w.p. \ } \ p=0.3 \\
      16, & \text{w.p. \ } \ p=0.2 \\
    \end{cases}
  \end{equation*}

The regime was designed to be centered around the mean value of $28$. As the original image size used for training is $32\times32$, we are now able to increase either the batch size or number of duplicates for each training step by a factor of  $\frac{32^2}{S^2}$ such that $S^2 \cdot B \cdot D$ is approximately constant.
We denote our modified mixed-size regimes as $B^+$ for an increased effective batch-size and $D^+$ for an increased number of BA duplicates of the same ratio.
We used our sampling strategy to train and compare our regime to the baseline results. We use the original hyper-parameters without modification. For the $B^+$  regime, use our gradient smoothing method, as described in Section \ref{sub:GradientSmoothing}. For each result, we measure our final test accuracy on the original $32 \times 32$ image size. We also perform batch-norm calibration as described in Section \ref{calibrate_bn}. 
From Table \ref{table:val_accuracy}, we see that our MixSize regimes on CIFAR datasets yield two possible improvements: 
\begin{itemize}
    \item Reduced number of training steps to achieve a similar test accuracy using $B^+$ regime.
    \item Better test accuracy when using $D^+$ regime.
\end{itemize}

\begin{figure}[t!]
\vspace{-2em}
    \centering
    \begin{subfigure}[b]{0.48\textwidth}
        \includegraphics[width=\textwidth,trim={0 0 0 0},clip]{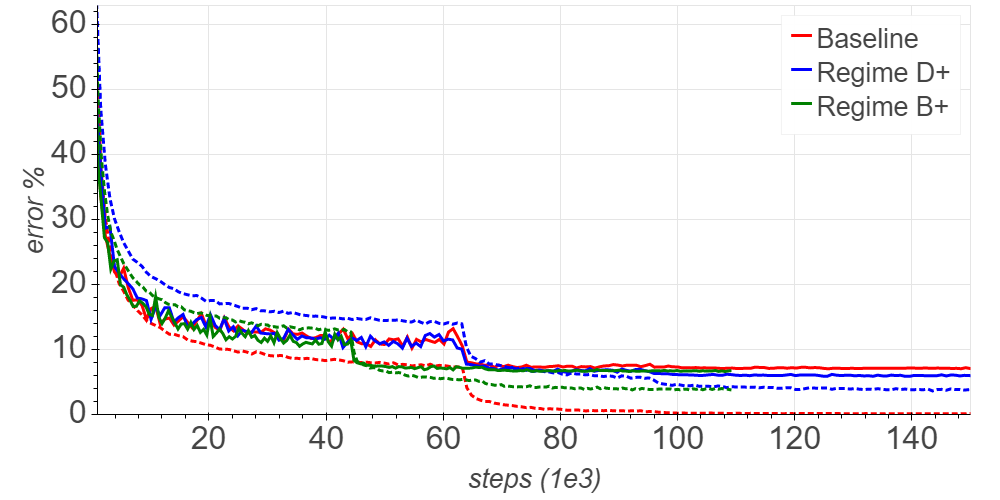}
        \caption{Test error}
        \label{compare:baseline}
    \end{subfigure}
    ~ 
    \begin{subfigure}[b]{0.48\textwidth}
        \includegraphics[width=\textwidth,trim={0 0 0 0},clip]{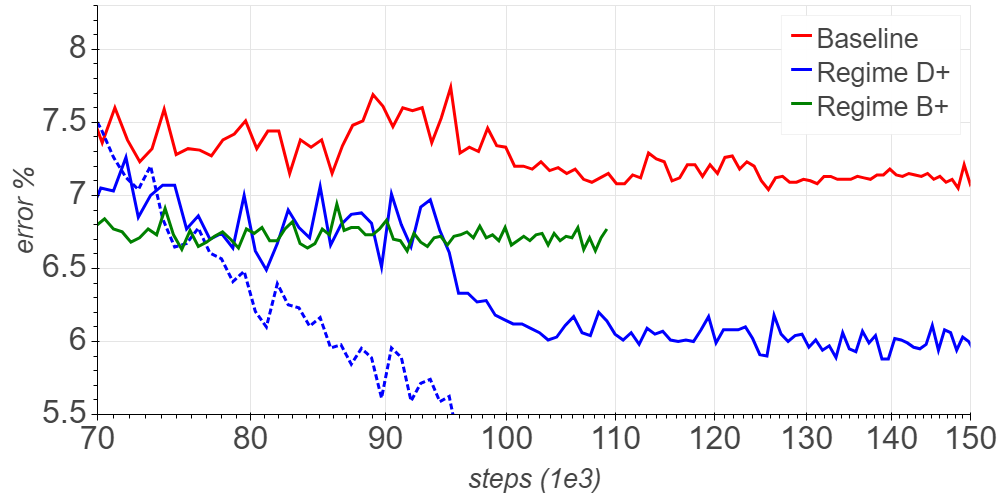}
        \caption{Final test error}
        \label{compare:zoom}
    \end{subfigure}
     \caption{Training (dotted) and test accuracy vs optimization step (ResNet44, CIFAR10). We compare vanilla training with two computationally equivalent stochastic regimes: increased duplicates ($D^+$) and increased batch ($B^+$). \textbf{$B^+$ regime achieves better test accuracy at a reduced number of iterations, while $D^+$ improves accuracy further at a similar computational cost}.}
    \label{compare_cifar}
    \vspace{-1em}
\end{figure}

Training progress on the CIFAR10 using ResNet44 is depicted in Figure \ref{compare_cifar}.
Interestingly, although designed only to reduce training time, we can see that our $B^+$ regime also improves accuracy in some cases. This improvement can be attributed to a regularization effect induced by changing image sizes during training, also manifested by an increase in training error throughout its progress. 
\begin{table*}[b]
\vspace{-1em}
\small
\centering{}
\caption{Test accuracy (Top-1) results for CIFAR and ImageNet. Each row represents models trained using the same computational and memory budget per step. Steps and accuracy are reported at the completion of a fixed epoch budget (e.g., $90$ epochs for ResNet on ImageNet, $200$ for ResNet on CIFAR). Accuracy is reported for model's original size ($32$ for CIFAR, $224$ for ImageNet).}
\label{table:val_accuracy}
\begin{tabular}{lccccccc}
\toprule{}\
      
Network                                        &      Dataset      &  \multicolumn{3}{c}{Steps}  & \multicolumn{3}{c}{Accuracy}    \\
\cmidrule(lr){3-5} 
\cmidrule(lr){6-8}    
                                               &                & Baseline  &  $B^+$ & $D^+$   & Baseline  &  $B^+$ & $D^+$\\

\midrule
ResNet-44                   &   CIFAR10 & 156K & \bf{109K} & 156K &  92.84\%    &  94.30\% & \textbf{94.37\%}  \\
WRN-28-10  &   CIFAR10   &  156K & \bf{109K} & 156K & 96.60\%       & 97.28\%  & \textbf{97.68\%}    \\
AmoebaNet     &   CIFAR10 &  469K & \bf{328K} & 469K & 98.16\%    &   98.14\%   & \textbf{98.32\%}    \\
\midrule
ResNet-44                     &                  CIFAR100 & 156K & \bf{109K} & 156K &  70.36\%   &  72.19\% & \textbf{73.10\%}    \\
WRN-28-10                    &           CIFAR100  & 156K & \bf{109K} & 156K & 79.85\%   &  83.08\%   & \textbf{83.52\%}    \\
\midrule
ResNet-50       &   ImageNet & 450K & \bf{169K} & 450K&  76.40\%   &   76.61\%    & \textbf{78.04\%} \\
EfficientNet-B0         &   ImageNet &  1000K & \bf{376K}  & 1000K & 76.32\% &  76.29\%     &  \textbf{76.53\%}             \\  
\bottomrule
\end{tabular}
\end{table*}

\vspace{-1.5em}
\paragraph{ImageNet.}
We also perform large scale experiments using the ImageNet dataset \citep{imagenet_cvpr09} to confirm our findings. We used the ResNet-50 \citep{he2016deep} model, with the training regime suggested by \citet{goyal2017accurate} that consists of base learning rate of $0.1$, decreased by a factor of $10$ on epochs $30,60,80$, stopping at epoch $90$. We used the base batch size of $256$ over $4$ devices and $L_2$ regularization over weights of convolutional layers. We used the standard data augmentation and did not incorporate any additional regularization or augmentation techniques.

Additionally, we also used the EfficientNet-B0 model suggested by \citet{tan2019efficientnet}. We used the same data augmentation and regularization as the original paper, but opted for a shorter training regime with a momentum-SGD optimizer that consisted of a cosine-annealed learning rate \citep{loshchilov2016sgdr} over $200$ epochs starting from an initial base $0.1$ value.

For the ImageNet dataset, we use the following stochastic regime found by cross-validation on several alternatives (see Appendix \ref{appendix:alt_size}):
  \begin{equation*}
    S^{(144)}: \  \ S=
    \begin{cases}
      256, & \text{w.p \ } \ p=0.1 \\
      224, & \text{w.p \ } \ p=0.1 \\
      128, & \text{w.p \ } \ p=0.6 \\
      96, & \text{w.p \ } \ p=0.2 \\
    \end{cases}
  \end{equation*}
  While the original training regime consisted of images of size $224 \times 224$, our proposed regime makes for an average image size of $\bar{S} \times \bar{S} = 144\times 144$. This regime was designed so that the reduced spatial size can be used to increase the corresponding batch size or the number of BA duplicates, as described in Section~\ref{stochastic_sizes}. We are first interested in accelerating the time needed for convergence of the tested models using our $B^+$ scheme. We enlarge the batch size used for each spatial size by a factor of $\frac{224^2}{S^2}$ such that $S^2 \cdot B$ is kept approximately fixed. 
  As the average batch size is larger than $B_o$, which was used with the original optimization hyper-parameters, we scale the learning rate linearly as suggested by \citet{goyal2017accurate} by a factor of $\frac{\bar{B}}{B_o}$. We note that for the proposed regimes we did not require any learning rate warm-up, due to the use of gradient smoothing.
  
  \begin{wrapfigure}{r}{0.5\textwidth}
\vspace{-1em}
  \begin{center}
    \includegraphics[width=.5\textwidth,trim={0 0 0 0cm},clip]{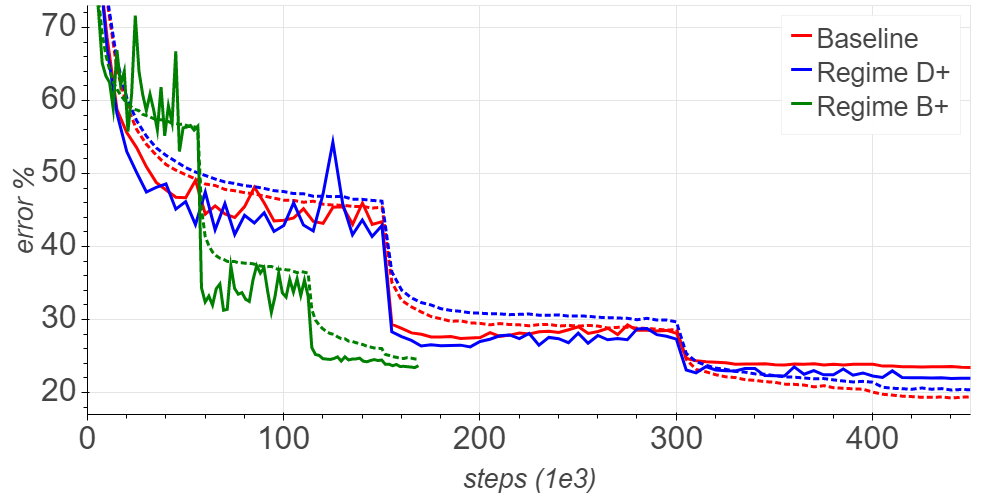}
    \caption{Training (dotted) and test accuracy on ImageNet using the Baseline, $B^+$ and $D^+$ regimes ($224\times 224$ evaluation size). All regimes required similar computational resources per step. $B^+$ regime required $\approx 2.7\times $ less steps per epoch. }\label{imagenet_regimes}
    \end{center}
    \vspace{-2em}
\end{wrapfigure} 
  As can be seen in Figure \ref{imagenet_regimes}, regime $B^+$ enables training with approximately $2.7\times$ less training steps, while reaching a better-than-baseline accuracy of $76.61\%$. As sizes were chosen to reflect in approximately equal computational cost per iteration, $B^{+}$ regime offers a similar improvement in total wall-clock time. 

  
  Next, we perform a similar experiment with a $D^+$ regime, where the number of BA duplicates is similarly increased with respect to $D_o$ instead of the batch size.
  This scaling results with an average duplicates of $\bar{D}=3$. 
  
  As the computational cost for each step remains approximately constant, as well as the number of required steps per epochs, training a model under this regime requires an equal wall-clock time. However, the increased batch-augmentation improves the final test accuracy to $78.04\%$, approximately $7\%$ relative improvement over the $76.4\%$ baseline.

\subsection{Increasing model resiliency to test-time changes in image size}
Next, we examine how MixSize affects the resulting model resiliency to changes in the image size during test-time. We evaluated the models by varying the test-time image sizes around the original $224$ spatial size: $S=224+ 32\cdot m, m\in \{-6,...,6\}$.
The common evaluation procedure for ImageNet models first scales the image to a $256$ smallest dimension and crops a center $224\times 224$ image. We adapt this regime for other image sizes by scaling the smallest dimension to $\floor{\frac{8}{7}S}$ (since $\frac{8}{7}\cdot224=256$) and then cropping the center $S\times S$ patch. 
Models trained with a mixed regime were calibrated to a specific evaluation size by measuring batch-norm statistics for $200$ batches of training samples. We note that for original fixed-size regimes this calibration procedure resulted with degraded results and so we report accuracy without calibration for these models. We did not use any fine-tuning procedure post training for any of the models. 

\begin{figure}[ht]
    \centering
    \begin{subfigure}[b]{0.48\textwidth}
    \centering
    \includegraphics[width=\textwidth,trim={0 0 0 0cm},clip]{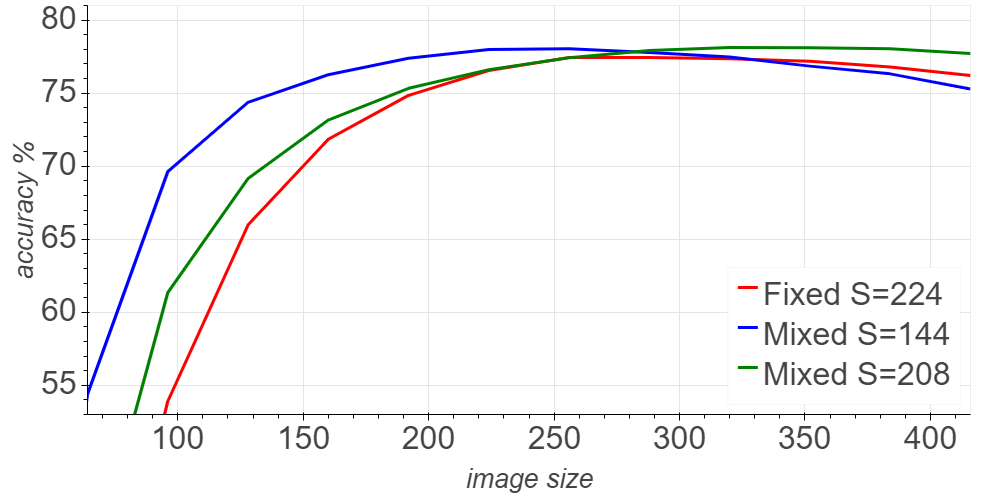}
    \caption{}
    \label{same_compute}
    \end{subfigure}
    ~ 
    \begin{subfigure}[b]{0.48\textwidth}
    \centering
    \includegraphics[width=\textwidth,trim={0 0 0 0cm},clip]{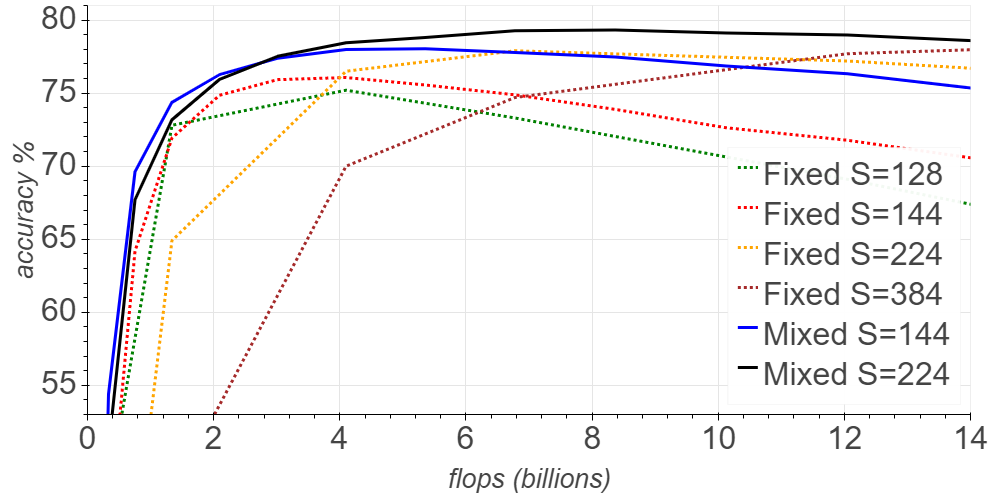}
    \caption{}
    \label{accuracy_per_flops_imagenet}
    \end{subfigure}
    \caption{Left: Test accuracy on validation set per image size, all models trained using the same computational and memory resources (regime $D^+$). Right: Test accuracy per billion flop 
    (at evaluation).}\label{compare_resnet50}
\end{figure}

As can be seen in Figure \ref{same_compute}, the baseline model trained using a fixed size, reaches $76.4\%$ top-1 accuracy at the same $224$ spatial size it was trained on. As observed previously, the model continues to slightly improve beyond that size, to a maximum of $76.8\%$ accuracy.
However, it is apparent that the model's performance quickly degrades when evaluating with sizes smaller than $224$. 

We compare these results with a $D^+$ regime, trained with an average size of $\bar{S}=144$. As described earlier, this model requires the same time and computational resources as the baseline model. However, due to the decreased average size, we were able to leverage more than $1$ duplicates per batch on average, which improved the model's top-1 accuracy to $77.14\%$ at size $224$. Furthermore, we find that the model performs much more favorably at image sizes smaller than $224$, scoring an improved (over baseline) accuracy of $76.43\%$ at only $160\times 160$ spatial size. 
We analyzed an alternative regime $S^{(208)}$, where the average spatial size is larger at $208\times 208$ (for more details see Appendix \ref{appendix:alt_size}). 
The model trained with the $S^{(208)}$ regime offers a similar improvement in accuracy, only across a larger spatial size, as it observed an average size of $208\times 208$ during training.
Figure \ref{same_compute} demonstrates that while all three models (Fixed with $S=224$, $S^{(144)}$ and $S^{(208)}$) were trained with the same compute and memory budget, mixed-size regimes offer superior accuracy over a wide range of evaluation sizes. Specifically, mixed-regime at $S=208$ dominates the baseline fixed-size regime at all sizes, while our mixed regime at $S=144$ achieves best results at sizes smaller than $224$. 

We also compared the classification performance across evaluated image sizes, using networks trained on a variety of fixed sizes and our mixed regimes. As a baseline, we use results obtained by \citet{fixresnet} (trained with repeated augmentations, without fine-tuning) and compare them with mixed-regime models trained with an equal computational budget, by setting the base number of BA duplicates to $D=2$. As can be seen in Figure \ref{accuracy_per_flops_imagenet}, mixed-regime trained models offer a wider range of resolutions with close-to-baseline accuracy (within a $2\%$ change) and perform better than their fixed-size counterparts at all sizes. As the number of floating-point operations (flops) grows linearly with the number of pixels, using a mixed regime significantly improves accuracy per compute at evaluation. We further note that our $S^{(224)}$ model reaches a top accuracy of $79.27\%$ at a $288 \times 288$ evaluation size.




\section{Summary}
In this work, we introduced and examined a performance trade-off between computational load and classification accuracy governed by the input's spatial size. We suggested stochastic image size regimes, which randomly change the spatial dimension as well as the batch size and the number of augmentation (duplicates) in the batch. Stochastic regime benefits are threefold: (1) reduced number of training iterations; or (2) improved model accuracy (generalization) and (3) improved model robustness to changing the image size. We believe this approach may have a profound impact on the practice of training convolutional networks. Given a computational and time budget, stochastic size regimes may enable to train networks faster, with better results, as well as to target specific image sizes that will be used at test time.  As the average size chosen to train is reflected in the optimal operating point for evaluation resolution, mixed regimes can be used to create networks with better performance across multiple designated use cases.
\bibliography{iclr2020_conference}
\bibliographystyle{iclr2020_conference}

\newpage
\appendix
\part*{Appendix}

\section{Experimental settings}
\subsection{CIFAR}
\label{experiments:cifar}
 We used the common data augmentation technique as described by \citet{he2016deep}. In this method, the input image is padded with $4$ zero-valued pixels at each side, top and bottom. A random $32 \times 32$ part of the padded image is then cropped and with a $0.5$ probability flipped horizontally. In order to adapt to varying input scales, we add an additional augmentation step, that resizes the images using bilinear interpolation  to $S \times S$, depending on a sampled size for each step. We note that this keeps the exact original augmentation procedure for $S=32$.

\section{Impact of gradient smoothing}
\begin{figure}[ht!]
    \centering
    \begin{subfigure}[b]{0.48\textwidth}
        \includegraphics[width=\textwidth,trim={0 0 0 0},clip]{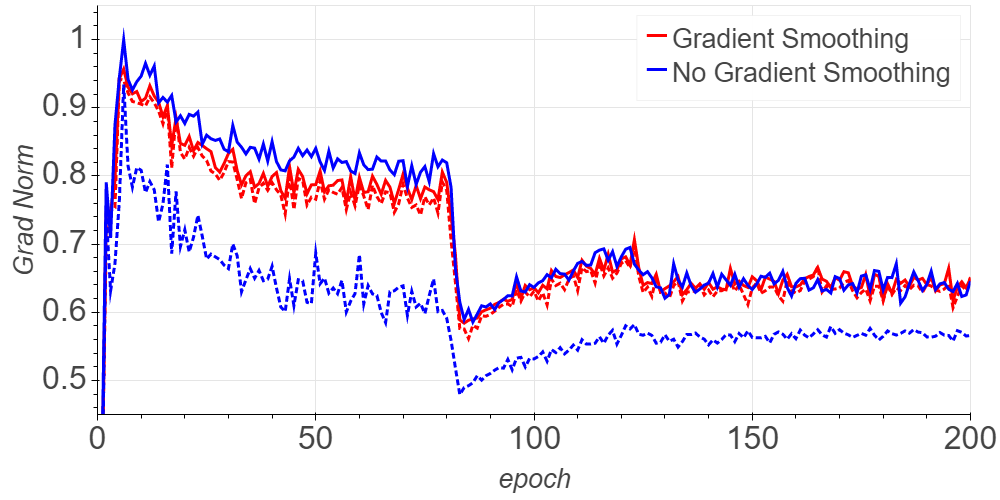}
        \caption{Gradient norm values with and without Grad-smoothing. Solid lines are gradients for $S=32$ while dotted lines are for $S=16$.}
        \label{gs_compare:grad}
    \end{subfigure}
    ~ 
    \begin{subfigure}[b]{0.48\textwidth}
        \includegraphics[width=\textwidth,trim={0 0 0 0},clip]{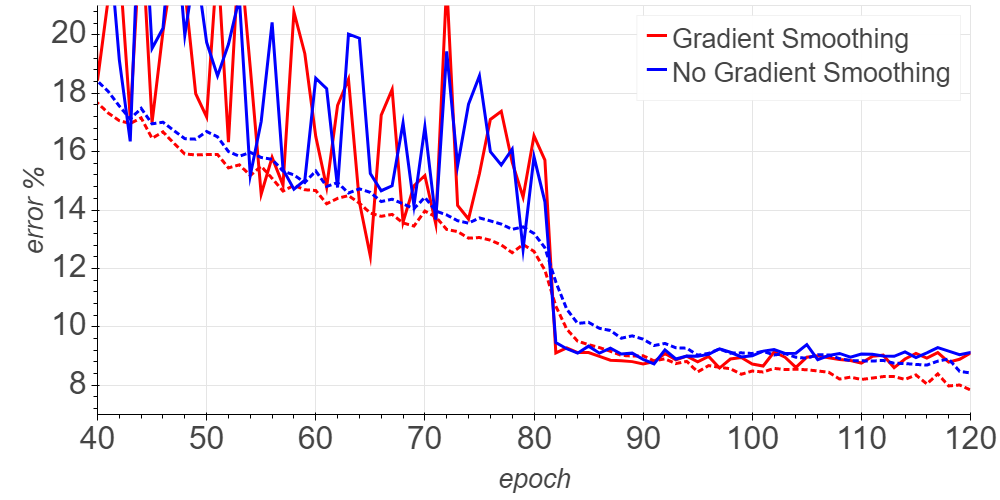}
        \caption{Training and test error with and with out gradient smoothing. Solid lines are test errors while dotted lines are for training.}
        \label{gs_compare:acc}
    \end{subfigure}
     \caption{Impact of gradient smoothing on CIFAR10, ResNet-44. The training regime includes two image sizes: $32\times 32$ and $16\times 16$ (average size is $S=24$). Using a $B^+$ regime creates two batch sizes: $256$ and $2,048$ respectively. Gradient smoothing helps to reduce gap between gradient norms at difference batch sizes and improves final accuracy.}
    \label{compare}
\end{figure} 

\section{Varying image-size training regimes}
We wish to consider training regimes with varying image sizes, such that the average image size is smaller than the desired evaluation size. For example, for the height dimension $H$, we wish to obtain an average size of $\bar{H}=\sum_i p_iH_i$ such that $\bar{H} < H_o$. We consider three alternatives for image size variations:
\begin{itemize}
    \item Increase image size from small to large, where each image size is used for number of epochs $E_i=p_iE_{\mathrm{total}}$, where $E_{\mathrm{total}}$ is the total number training epochs required.
    \item Using a random image size for each epoch, keeping the epoch number for each size at $E_i$
    \item Sampling image size per training step at probability $p_i$
\end{itemize}

\begin{figure}[ht!]
    \centering
    \begin{subfigure}[b]{0.48\textwidth}
        \includegraphics[width=\textwidth,trim={0 0 0 0},clip]{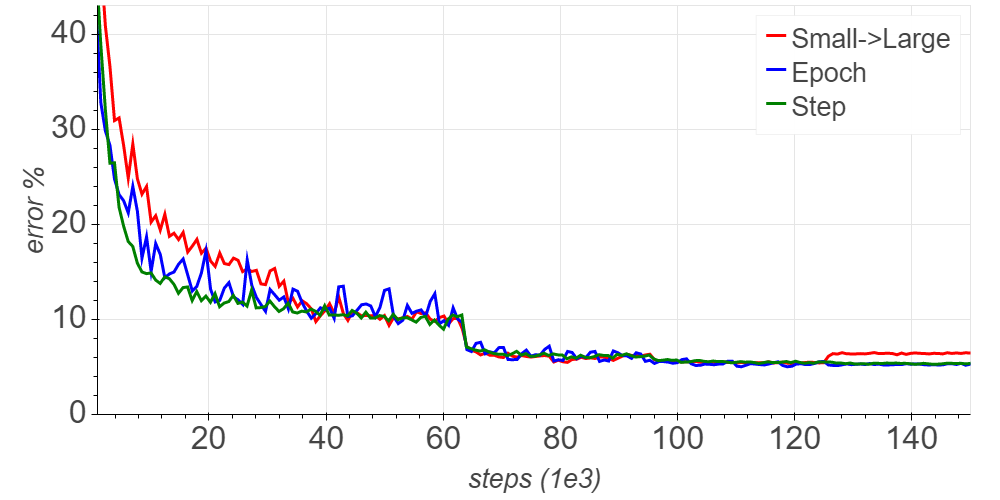}
        \caption{Test error}
        \label{compare:baseline}
    \end{subfigure}
    ~ 
    \begin{subfigure}[b]{0.48\textwidth}
        \includegraphics[width=\textwidth,trim={0 0 0 0},clip]{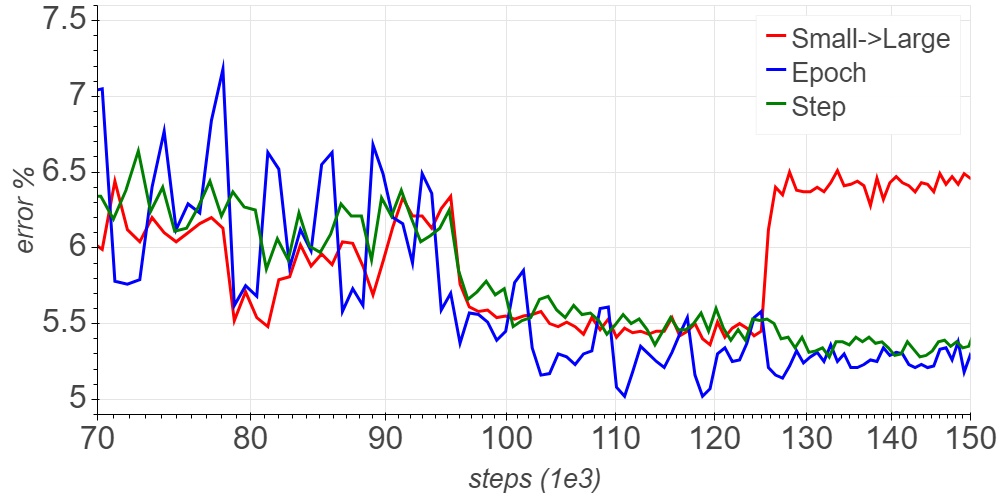}
        \caption{Final test error}
        \label{compare:zoom}
    \end{subfigure}
     \caption{Test accuracy vs step for 3 size sampling regimes: (1) From small to large (2) Sample each Epoch (3) Sample each step. All methods reached a similar accuracy, but sampling each epoch was less noisy and did not require hyper-parameter tuning.}
    \label{size_regimes}
\end{figure}

As can be seen in Figure \ref{size_regimes}, we found that random sampling regimes performed better than scaling image size from small to large \citep{FastAi, fixresnet}. While sampling both at epoch and step time frames performed similarly, replacing sizes on each step seemed to converge faster and to have less noise in measured test accuracy. We note that these behaviours may partly stem from the use of batch-normalization \citep{ioffe2015batch} which is sensitive to the image size used at evaluation or insufficient hyper-parameter tuning for each specific size (e.g., spiking error at the end of the small-to-large regime). Considering these findings, we continue to perform our experiments using the third regime -- sampling image size per training step. 

\section{Alternative size distribution regimes}
\label{appendix:alt_size}
We used alternative size regimes balanced around $224$, named $S^{(208)}$ and  $S^{(224)}$. They can be described by the following distributions: 
  \begin{equation*}
    S^{(208)}: \  \ S=
    \begin{cases}
      320, & \text{w.p \ } \ p=0.1 \\
      288, & \text{w.p \ } \ p=0.1 \\
      256, & \text{w.p \ } \ p=0.1 \\
      224, & \text{w.p \ } \ p=0.2 \\
      192, & \text{w.p \ } \ p=0.2 \\
      160, & \text{w.p \ } \ p=0.1 \\
      128, & \text{w.p \ } \ p=0.1 \\
      96, & \text{w.p \ } \ p=0.1 \\
    \end{cases}
  \end{equation*}
  
    \begin{equation*}
    S^{(224)}: \  \ S=
    \begin{cases}
      320, & \text{w.p \ } \ p=0.133 \\
      288, & \text{w.p \ } \ p=0.133 \\
      256, & \text{w.p \ } \ p=0.133 \\
      224, & \text{w.p \ } \ p=0.2 \\
      192, & \text{w.p \ } \ p=0.133 \\
      160, & \text{w.p \ } \ p=0.133 \\
      128, & \text{w.p \ } \ p=0.133 \\
    \end{cases}
  \end{equation*}
\end{document}